\documentclass{article}

 \PassOptionsToPackage{numbers, compress}{natbib}

 \usepackage[final]{nips_2018}

\usepackage{microtype}
\usepackage{graphicx}
\usepackage{subfigure}
\usepackage{booktabs} 

\newcommand\underrel[2]{\mathrel{\mathop{#2}\limits_{#1}}}
\usepackage{bbm}
\usepackage[T1]{fontenc}    
\usepackage{hyperref}       
\usepackage{url}            
\usepackage{booktabs}       
\usepackage{amsfonts,amsmath,amssymb,stmaryrd}
\usepackage{nicefrac}       
\usepackage{microtype}      
\usepackage{algorithm} 
\usepackage{algorithmic} 

\title{Cluster Variational Approximations for Structure Learning of Continuous-Time Bayesian Networks from Incomplete Data}

\author{
  Dominik~Linzner\thanks{This research is funded by the European Union's Horizon 2020
research and innovation programme under grant agreement
668858.} \\
  Department of Electrical Engineering and Information Technology\\
  Technische Universität Darmstadt\\
  \texttt{dominik.linzner@bcs.tu-darmstadt.de} \\
  \And
  Heinz~Koeppl$^*$ \\
  Department of Electrical Engineering and Information Technology\\
  Department of Biology\\
  Technische Universität Darmstadt\\
  \texttt{heinz.koeppl@bcs.tu-darmstadt.de} \\
}

\begin{document}

\maketitle

\begin{abstract}
Continuous-time Bayesian networks (CTBNs) constitute a general and powerful framework for modeling continuous-time stochastic processes on networks. This makes them particularly attractive for learning the directed structures among interacting entities. However, if the available data is incomplete, one needs to simulate the prohibitively complex CTBN dynamics. Existing approximation techniques, such as sampling and low-order variational methods, either scale unfavorably in system size, or are unsatisfactory in terms of accuracy. Inspired by recent advances in statistical physics, we present a new approximation scheme based on cluster-variational methods significantly improving upon existing variational approximations. We can analytically marginalize the parameters of the approximate CTBN, as these are of secondary importance for structure learning. This recovers a scalable scheme for direct structure learning from incomplete and noisy time-series data. Our approach outperforms existing methods in terms of scalability.
\end{abstract}

\section{Introduction}
Learning directed structures among multiple entities from data is an important problem with broad applicability, especially in biological sciences, such as genomics \cite{Acerbi2014} or neuroscience \cite{Schadt2005}. 
With prevalent methods of high-throughput biology, thousands of molecular components can be monitored simultaneously in abundance and time.
Changes of biological processes can be modeled as transitions of a latent state,  such as expression or non-expression of a gene or activation/ inactivation of protein activity. However, processes at the bio-molecular level evolve across vastly different time-scales \cite{Klann2012}. Hence tracking every transition between states is unrealistic. Additionally, biological systems are, in general, strongly corrupted by measurement or intrinsic noise.

In previous numerical studies, continuous-time Bayesian networks (CTBNs) \cite{Nodelman1995} have been shown to outperform competing methods for reconstruction of directed networks, such as ones based on Granger causality or the closely related dynamic Bayesian networks  \cite{Acerbi2014}. Yet, CTBNs suffer from the curse of dimensionality, prevalent in multi-component systems. This becomes problematic if observations are incomplete, as then the latent state of a CTBN has to be laboriously estimated \cite{Nodelman2005}.
In order to tackle this problem, approximation methods through sampling, e.g., \cite{Fan2008,El-Hay2011,Rao2012}, or variational approaches \cite{Cohn2010,El-Hay2010} have been investigated. These, however, either fail to treat high-dimensional spaces because of sample sparsity, are unsatisfactory in terms of accuracy, or provide good accuracy at the cost of an only locally consistent description.



In this manuscript, we present, to the best of our knowledge, the first direct structure learning method for CTBNs based on variational inference.
Our method combines two key ideas. We extend the framework of variational inference for multi-component Markov chains by borrowing results from statistical physics on cluster-variational-methods \cite{Yedidia2000,Vazquez2017,Pelizzola2017}. Here the previous result in \cite{Cohn2010} is recovered as a special case. We show how to calculate parameter-free dynamics of CTBNs in form of ordinary differential equations (ODEs), depending only on the observations, prior assumptions, and the graph structure. Lastly, we derive an approximation for the structure score, which we use to implement a scalable structure learning algorithm. 
The notion of using marginal CTBN dynamics for network reconstruction from noisy and incomplete observations was recently explored in \cite{Studer2016} to successfully reconstruct networks of up to 11 nodes by sampling from the exact marginal posterior of the process, albeit using large computational effort. Yet, the method is sampling-based and thus still scales unfavorably in high dimensions.
In contrast, we can recover the marginal CTBN dynamics at once using a standard ODE solver.


\section{Background}
\subsection{Continuous-time Bayesian networks}
We consider continuous-time Markov chains (CTMCs) $\{X(t)\}_{t\geq 0}$ taking values in a countable state-space $\mathcal{X}$. A time-homogeneous Markov chain evolves according to an intensity matrix $\mathcal{R}:\mathcal{X}\times\mathcal{X}\rightarrow  \mathbb{R}$, whose elements are denoted by $\mathcal{R}(x,y)$, where  $x, y\in\mathcal{X}$. 

 A continuous-time Bayesian network \cite{Nodelman1995} is defined as an $N$-component process over a factorized state-space $\mathcal{X}=\mathcal{X}_1\times\dots\times\mathcal{X}_N$ evolving jointly as a CTMC. In the following, we will make use of the shorthand $x=(x_1,\dots,x_N)$ with $x\in\mathcal{X}$ and  $x_n\in\mathcal{X}_n$. However, as mostly no ambiguity arises we write $x$ for $x\in\mathcal{X}_n$ in these cases to lighten the notation. We impose a directed graph structure $\mathcal{G}=(V,E)$, encoding the relationship among the components $V\equiv\{V_1,\dots,V_N\}$, which we refer to as nodes. These are connected via an edge set $E\subseteq V\times V$. This quantity -- the structure -- is what we will later learn. The instantaneous state of each component is denoted by $X_n(t)$ assuming values in $\mathcal{X}_n$, which depends only on the states of a subset of nodes, called the parent set $\mathrm{pa}(n)\equiv\{m \mid  (m,n)\in E\}$. Conversely, we define the child set $\mathrm{ch}(n)\equiv\{m \mid  (n,m)\in E\}$. The dynamics of a local state $X_n(t)$ are modeled as a Markov process, when conditioned on the current state of all its parents ${U}_n(t)$ taking values in $\mathcal{U}_n\equiv\{\mathcal{X}_m\mid m\in\mathrm{pa}(n)\}$. They can then be expressed by means of the conditional intensity matrices (CIMs) $\mathcal{R}_{n}^u:\mathcal{X}_n\times\mathcal{X}_n\rightarrow  \mathbb{R}$, where $u\equiv(u_1,\dots u_{L})\in\mathcal{U}_n$ denotes the current state of the parents ($L=|\mathrm{pa}(n)|$).
Specifically, we can express the probability of finding node $n$ in state $y$ after some small time-step $\delta t$, given that it was in state $x$ at time $t$ with $x,y\in\mathcal{X}_n$ as
\begin{align*}
 P(X_n(t+{\delta t})=y\mid X_n(t)=x,U_n(t)=u)=\delta(x,y)+\mathcal{R}_{n}^u(x,y){\delta t}+{o}({\delta t}),
\end{align*}
where $\mathcal{R}_{n}^u(x,y)$ is the matrix element of $ \mathcal{\mathcal{R}}_{n}^u$ corresponding to the transition $x\rightarrow y$ given the parents' state $u$. Additionally, we have to enforce $\mathcal{R}_{n}^u(x,x)\equiv-\sum_{y\neq x}\mathcal{R}_{n}^u(x,y)$ for $ \mathcal{\mathcal{R}}_{n}^u$ to be a proper CIM. The CIMs are connected to the joint intensity matrix $\mathcal{R}$ of the CTMC via amalgamation -- see, for example, \cite{Nodelman1995}.

\subsection{Variational energy}
The foundation of this work is to derive a lower bound on the evidence of the data for a CTMC in the form of a variational energy. Such variational lower bounds are of great practical significance and pave the way to a multitude of approximate inference methods, in this context called \emph{variational inference}. We consider paths $\bold{X}\equiv\{X(s)\mid 0 \leq s \leq T\}$ of a CTMC  with a series of noisy state observations $\bold{Y}\equiv(Y^0,\dots,Y^{I})$ at times $(t^0,\dots,t^{I})$, drawn according to an observation model $Y^i\sim P(Y^i\mid X(t^i))$. 
We consider the posterior Kullback--Leibler (KL) divergence $D_{KL}(Q(\bold{X})| | P(\bold{X}\mid \bold{Y}))$ given a candidate distribution $Q(\bold{X})$, which can be decomposed as
\begin{align*}
D_{KL}(Q(\bold{X})| | P(\bold{X} \mid \bold{Y}))=&  D_{KL}(Q(\bold{X})| | P(\bold{X})P(\bold{Y}\mid \bold{X}))+\ln P(\bold{Y}).
\end{align*}
As $D_{KL}(Q(\bold{X})| | P(\bold{X} \mid \bold{Y}))\geq0$ this recovers a lower bound on the evidence
\begin{align}
\ln P(\bold{Y})\geq \mathcal{F},
\end{align}
where the bound $\mathcal{{F}}\equiv - D_{KL}(Q(\bold{X})| | P(\bold{X})P(\bold{Y}\mid\bold{X}))$ is also known as the Kikuchi functional \cite{Kikuchi1951}, or the Kikuchi variational energy. The Kikuchi functional has recently found heavy use in variational approximations for probabilistic models \cite{Yedidia2000,Vazquez2017,Pelizzola2017}, because of the freedom it provides for choosing clusters in space and time. We will now make use of this feature.

%
\section{Cluster variational approximations for CTBNs}
\begin{figure}[t]
	\begin{center}
		\includegraphics[width=0.3\columnwidth]{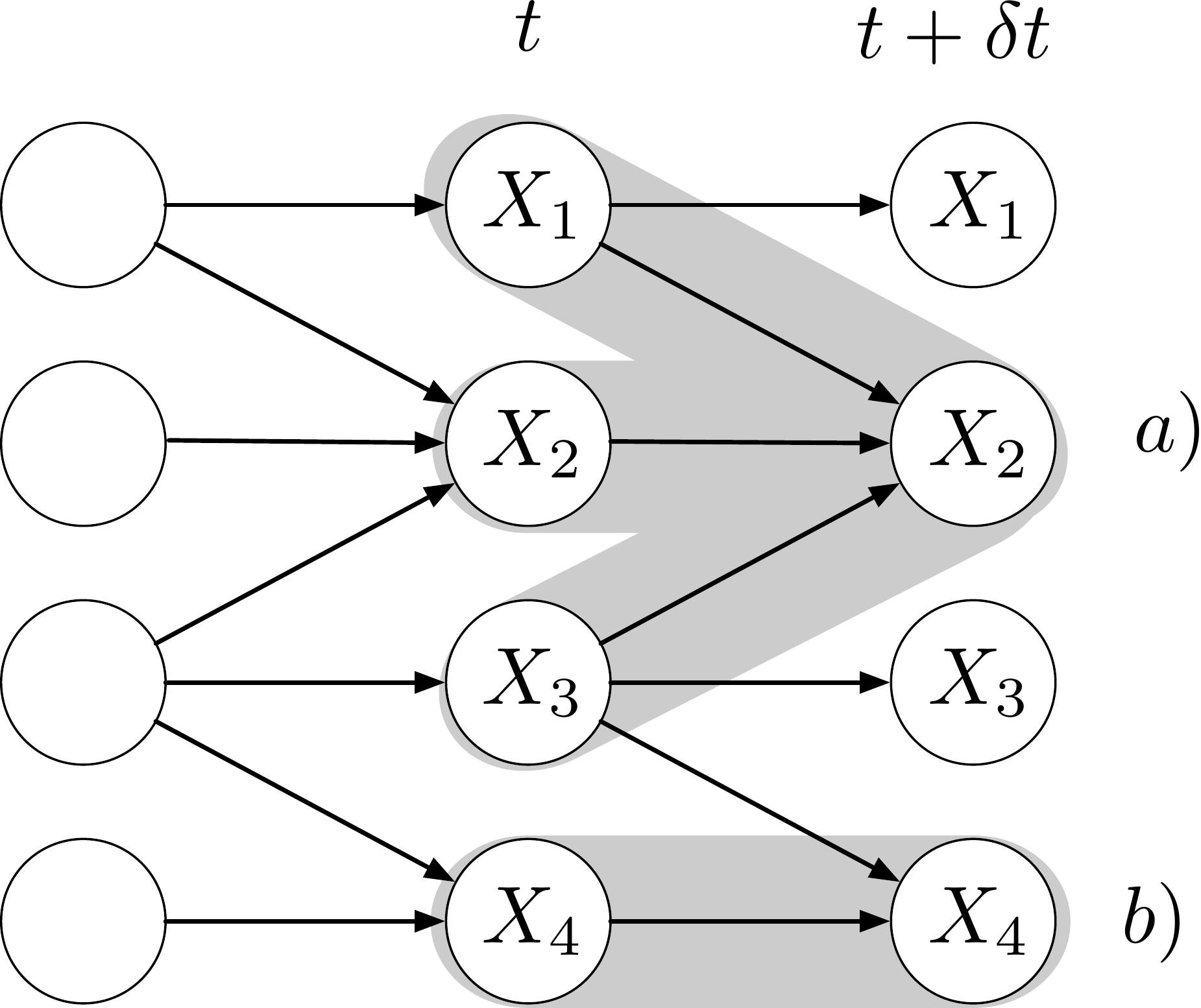}
		\end{center}
	\caption{Sketch of different cluster choices for a CTBN in discretized time : a) star approximation b) naive mean-field.}
	\label{clusters}
\end{figure}
The idea behind cluster variational approximations, derived subsequently, is to find a decomposition of the Kikuchi functional over $M$ cluster functionals $\mathcal{F}_j$ of smaller sub-graphs $A_j(t)$ for a CTBN using its $\delta t$-discretization (see Figure \ref{clusters}):
\begin{align*}
\mathcal{F}\simeq \int_{0}^{T}\mathrm{d}t \sum_{j=1}^M \mathcal{F}_j(A_j(t)).
\end{align*}
Examples for $A_j(t)$ are the the completely local \emph{naive mean-field approximation} $A^\mathrm{mf}_j(t)=\{X_j(t+\delta t),X_j(t)\}$, or the \emph{star approximation} $A^\mathrm{s}_j(t)=\{X_j(t+\delta t),{U}_j(t),X_j(t)\}$ on which our method is based. 
We notice that the formulation of CTBNs already imposes structure on the transition matrix
\begin{align}\label{eq:CTBN-semantics}
P(X(t+\delta t)\mid X(t))=\prod_{n=1}^N P(X_n(t+\delta t)\mid X_n(t),{U}_n(t)),
\end{align}
suggesting a node-wise factorization to be a natural choice. Our goal is to find an expansion of $\mathcal{F}$ for finite $\delta t$ for different cluster choices and subsequently consider the continuous-time limit $\delta t\rightarrow 0$.
In order to arrive at the Kikuchi functional in the star approximation, we assume that $Q(\bold{X})$ describes a CTBN, i.e. satisfies \eqref{eq:CTBN-semantics}. However, to render our approximation tractable, we further restrict the set of approximating processes by assuming them to be only weakly dependent in the transitions. Specifically, we require the existence of some expansion in orders of the \emph{coupling strength} $\varepsilon$
\begin{align}
 \notag Q(X_n(t+\delta t)\mid X_n(t),{U}_n(t))= Q(X_n(t+\delta t)\mid X_n(t))+\mathcal{O}(\varepsilon),
\end{align}
where the remainder $\mathcal{O}(\varepsilon)$ contains the dependency on the parents.\footnote{An example of a function with such an expansion is a Markov random field with coupling strength $\varepsilon$.}
Because the derivation is quite lengthy, we have to leave the details of the calculation to Appendix B.1.
The Kikuchi functional $\mathcal{F}$ can then be expanded in first order of $\varepsilon$ and decomposes on the $\delta t$-discretized network spanned by the CTBN process, into local star-shaped terms -- see, for example, Figure \ref{clusters}.
We emphasize that the expanded variational energy in star approximation is no longer a lower bound on the evidence, but provides an approximation.
We define a set of marginals, completely specifying a CTBN
\begin{align*}
&{m}_n(x)\equiv Q(X_n(t)=x),\\
&\tau^u_n(x,y)\equiv \lim_{\delta t\rightarrow 0}\frac{Q(X_n(t+\delta t)=y,X_n(t)=x,{U}_n(t)=u)}{\delta t}\quad \text{ for } x\neq y,
\end{align*}
the shorthand ${m_n^u}\equiv \prod_{l\in\mathrm{pa}(n)}{m}_l(u_l)$ and $\tau^u_n(x,x)\equiv-\sum_{y\neq x}\tau^u_n(x,y)$.
Checking self-consistency of these quantities via marginalization of $Q(X_n(t+\delta t)=y,X_n(t)=x,{U}_n(t)=u)$ recovers an inhomogeneous Master equation 
\begin{align}
\dot{m}_n(x)=\sum_{y\neq x,u}[\tau^u_n(y,x)-\tau^u_n(x,y)].\label{eq:continuity}
\end{align}
Because of the intrinsic asynchronous update constraint on CTBNs, only local probability flow inside the state-space $\mathcal{X}_n$ is allowed. This renders this equation equivalent to a continuity constraint on the global probability distribution. 
The resulting functional is only dependent on the marginal distributions. Performing the limit of $\delta t\rightarrow 0$, we arrive at a sum of node-wise functionals in continuous-time (see Appendix B.2)
\begin{align*}
\mathcal{{F}}\simeq\mathcal{{F}}_{S}, \quad \mathcal{{F}}_{S}\equiv\sum_{n=1}^N (H_{n}+E_{n})+\mathcal{F}_0,
\end{align*}
where we identified the entropy $H_n$ and the energy  $E_n$ respectively
\begin{align*}
H_n&=\int_0^T \mathrm{d}t \sum_{x,u}\sum_{y\neq x}\tau^u_n(x,y)\left[1-\ln\tau^u_n(x,y)+\ln(m_n(x)m_n^u)\right], \\
E_n&=\int_0^T \mathrm{d}t\left[\sum_{x}m_n(x)\mathbb{E}_{{n}}[\mathcal{R}_{n}^u(x,x)] +\sum_{x,u}\sum_{y\neq x}\tau^u_n(x,y)\ln \mathcal{R}_{n}^u(x,y)\right].
\end{align*}
The expectation value is defined as $\mathbb{E}_{n}[f(u)]\equiv\sum_{u'}m_n^{u'} f(u')$ for any function $f(u)$.
As expected, the functional is very similar to the one derived in \cite{Cohn2010} as both are derived from the KL divergence between true and approximate distributions from a set of marginals. Indeed, if we replace our star-shape cluster by the completely local one $A^{\mathrm{mf}}_j(t)$, we recover exactly their previous result, demonstrating the generality of our method (see Appendix B.3). 
In principle, higher-order clusters can be considered \cite{Vazquez2017,Pelizzola2017}. Lastly, we enforce continuity by \eqref{eq:continuity} fulfilling the constraint. We can then derive the Euler-Lagrange equations corresponding to the Lagrangian,
\begin{align*}
\mathcal{{L}}=
\mathcal{{F}}-\int_0^T \mathrm{d}t\sum_{n,x}\lambda_n(x)\left\{\dot{m}_n(x)-\sum_{y\neq x,u}[\tau^u_n(y,x)-\tau^u_n(x,y)]\right\},
\end{align*}  
with Lagrange multipliers $\lambda_n(x)\equiv\lambda_n(x,t)$. 


\subsection{CTBN dynamics in star approximation}
\begin{figure*}[t]
\vskip 0.2in
\begin{center}
\centerline{\includegraphics[width=0.85\columnwidth]{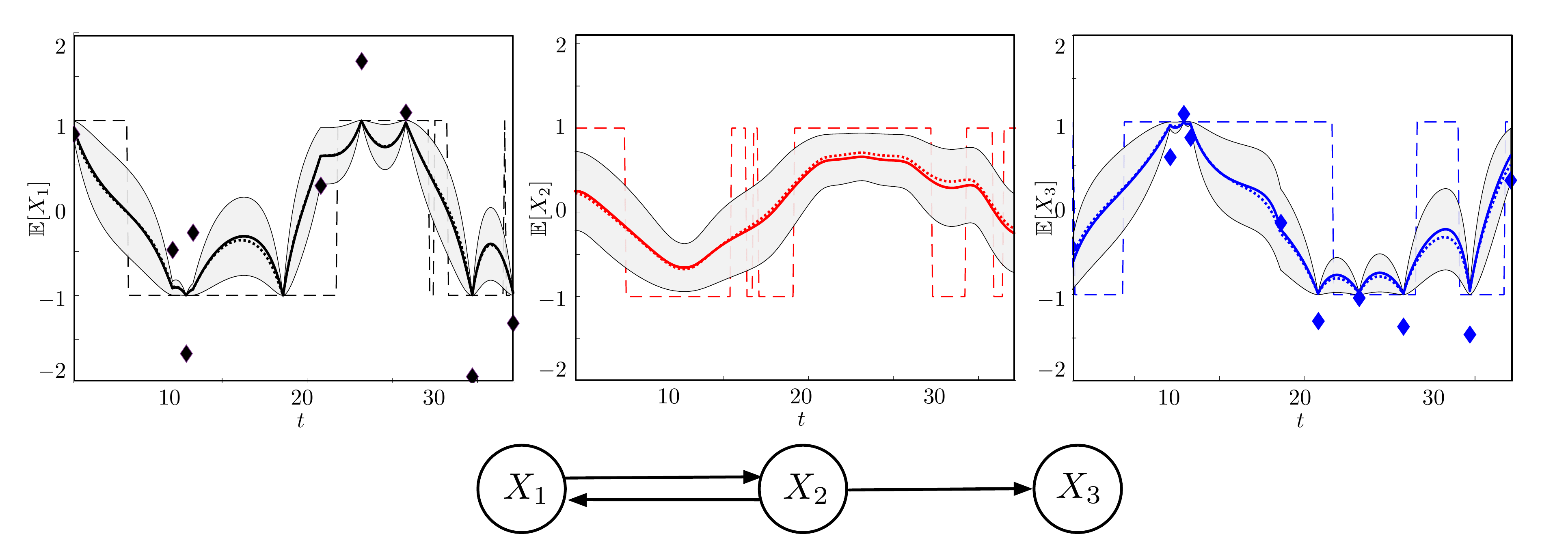}}
\caption{Dynamics in star approximation of a 3 node CTBN following Glauber dynamics at $a=1$ and $b=0.6$ conditioned on noisy observations (diamonds). We plotted the expected state (blue) plus variance (grey area). The observation model is the latent state plus gaussian random noise of variance $\sigma=0.8$ and zero mean. The latent state (dashed) is well estimated for $X_2$, even when no data has been provided. For comparison, we plotted the exact posterior mean (dots). We did not plot the exact variance, which depends only on the mean, for better visibility.}
\label{NOISY_CTBN}
\end{center}
\vskip -0.2in
\end{figure*}

The approximate dynamics of the CTBN can be recovered as stationary points of the Lagrangian, satisfying the Euler--Lagrange equation.
Differentiating $\mathcal{{L}}$ with respect to ${m}_n(x)$, its time-derivative $\dot{m}_n(x)$, $\tau^u_n(x,y)$ and the Lagrange multiplier $\lambda_n(x)$ yield a closed set of coupled 
ODEs for the posterior process of the marginal distributions ${m}_n(x)$ and transformed Lagrange multipliers $\rho_n(x)\equiv\exp(\lambda_n(x))$, eliminating $\tau^u_n(x,y)$
\begin{align}
\label{eq:ode-back} \dot{\rho}_n(x)=&\{\mathbb{E}_n\left[{\mathcal{R}}_{n}^u(x,x)\right]+\psi_n(x) \}\rho_n(x)-\sum_{y\neq x}\mathbb{E}_n\left[{\mathcal{R}}_{n}^u(x,y)\right]\rho_n(y), \\
\label{eq:ode-forw} \dot{m}_n(x)=& \sum_{y\neq x}m_n(y)\mathbb{E}_n[{\mathcal{R}}_{n}^u(y,x)]\frac{\rho_n(x)}{\rho_n(y)} -m_n(x)\mathbb{E}_n[{\mathcal{R}}_{n}^u(x,y)]\frac{\rho_n(y)}{\rho_n(x)},
\end{align} 
with  
\begin{align*}
&\psi_n(x')=\sum_{j\in\mathrm{ch}(n)}\sum_{x}m_j(x)\left\{\sum_{y\neq x}\frac{\rho_j(y)}{\rho_j(x)} \mathbb{E}_j[{\mathcal{R}}_{j}^u(x,y)|X_n(t)=x']+\mathbb{E}_j[{\mathcal{R}}_{j}^u(x,x)|X_n(t)=x']\right\}.
\end{align*} 
Furthermore, we recover the \emph{reset condition} 
\begin{align}\label{eq:reset}
\lim_{t\rightarrow t^{i-}}\rho_n(x)=\lim_{t\rightarrow t^{i+}}\rho_n(x)\exp\left\{\sum_{x'\in\mathcal{X}\mid x'_n=x}\ln P(Y^i\mid x')\prod_{k=1,k\neq n}^N m_k(x'_k)\right\},
\end{align}
and $x_k$ denotes the $k$th component of $x\in\mathcal{X}$. This incorporates the conditioning of the dynamics on noisy observations. For the full derivation we refer the reader to Appendix B.4. 
We require boundary conditions for the evolution interval in order to determine a unique solution to the set of equations \eqref{eq:ode-back} and \eqref{eq:ode-forw}. 
We thus set either $m_n(x,0)=Y_n^0$ and $\rho_n(x,T)=Y_n^I$ in the case of noiseless observations, or -- if the observations have been corrupted by noise -- $m_n(x,0)=\frac{1}{2}$ and $\rho_n(x,T)=1$ as boundaries before and after the first and the last observation, respectively. The coupled set of ODEs can then be solved iteratively as a fixed-point procedure in the same manner as in previous works \cite{Opper2008,Cohn2010} (see Appendix A.1 for details) in a forward-backward procedure. As the Kikuchi functional is convex, this procedure is guaranteed to converge.
In order to incorporate noisy observations into the CTBN dynamics, we need to assume an observation model. In the following we assume that the data likelihood factorizes $
P(Y^i\mid X)=\prod_n P(Y_n^i\mid X_n)$,
allowing us to condition on the data by enforcing $\lim_{t\rightarrow t^{i-}}\rho_n(x)=\lim_{t\rightarrow t^{i+}} P_n(Y^i\mid x)\rho_n(x)$. In Figure \ref{NOISY_CTBN}, we exemplify CTBN dynamics $(N=3)$ conditioned on observations corrupted by independent Gaussian noise. We find close agreement with the exact posterior dynamics. Because we only need to solve $2N$ ODEs to approximate the dynamics of an $N$-component system, we recover a linear complexity in the number of components, rendering our method scalable.

\subsection{Parameter estimation}
Maximization of the variational energy with respect to transition rates ${\mathcal{R}}_{n}^u(x,y)$ yields the expected result for the estimator of transition rates 
\begin{align*}
\hat{\mathcal{R}}_{n}^u(x,y)=\frac{\mathbb{E}[M^u_n(x,y)]}{\mathbb{E}[T^u_n(x)]},
\end{align*}
given the \emph{expected sufficient statistics} \cite{Nodelman2005}:
\begin{align*}
\mathbb{E}[T^u_n(x)]=\int_0^T\mathrm{d}t \ m_n(x)m_n^u, \quad \mathbb{E}[M^u_n(x,y)]=\int_0^T\mathrm{d}t \ \tau^u_n(x,y),
\end{align*}
where $\mathbb{E}[T^u_n(x)]$ are the \emph{expected dwelling times} and $\mathbb{E}[M^u_n(x,y)]$ are the \emph{expected number of transitions}.
Following standard  \emph{expectation--maximization} (EM) procedure, e.g. \cite{Opper2008}, we can estimate the systems' parameters given the underlying network.
 \begin{figure}[ht]
	\begin{center}
		\includegraphics[width=1.00\columnwidth]{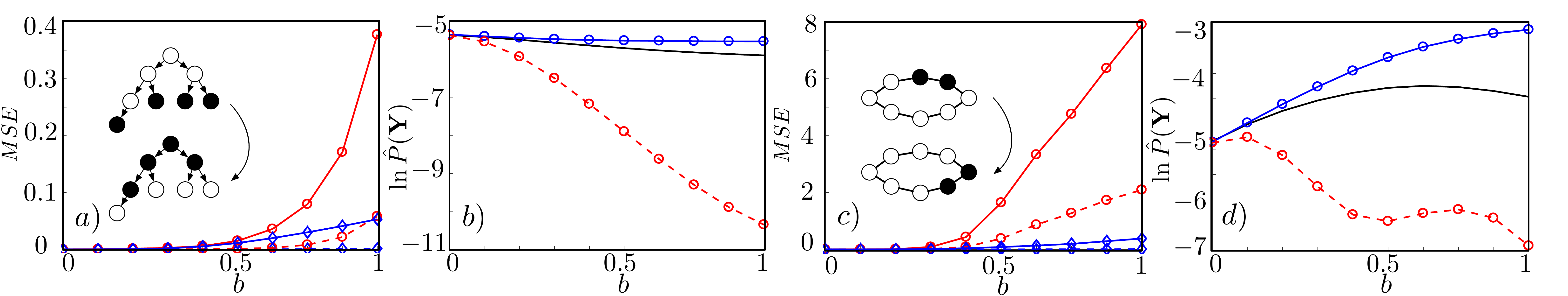}
		\end{center}
	\caption{We perform inference on a tree network a) and b), and an undirected chain c) and (d), both consisting of 8 nodes with noiseless evidence as denoted in sketch inlet (black: $x=-1$, white: $x=1$) in a) and c) obeying Glauber dynamics with $a=8$. In a) we plotted mean-squared-error ($\mathrm{MSE}$) for the expected dwelling times (dashed) and the the expected number of transitions for the naive mean-field (circle, red) and star approximation (diamond, blue) with respect to the predictions of the exact simulation as a function of temperature $b$. In b) and d) we plot the approximation of logarithmic evidence as a function of temperature. We find that for both approximations (star approximation in blue, naive mean field in red dashed and exact result in black) better performance on the tree network, while the star approximation clearly improves upon naive mean-field in both scenarios.}
	\label{suff_stat}
\end{figure}

\subsection{Benchmark}
In the following we compare the accuracy of the star approximation with the naive mean-field approximation. Throughout this section we will consider a binary local state-space (spins) $\mathcal{X}_n=\{+1,-1\}$. We consider a system obeying Glauber dynamics with the rates $
{\mathcal{R}}_{n}^u(x,-x)=\frac{a}{2} \left( 1+x\tanh\left( b\sum_{l\in\mathrm{pa}(n)}u_l\right)\right).
$
Here $b$ is the inverse temperature of the system. With increasing $b$ the dynamics of each node depend more strongly on the dynamics of its neighbors and thus harder to describe using mean-field dynamics. The pre-factor $a$ scales the overall rate of the process. This system can be an appropriate toy-example for biological networks as it encodes additive threshold behavior. In Figure \ref{suff_stat} $a)$ and $c)$, we show how closely the expected sufficient statistics match the true ones for a tree network and an undirected chain with periodic boundaries of 8 nodes, so that we can still compare to the exact result. In this application, we restrict ourselves to noiseless observations to better connect to previous results as in \cite{Cohn2010}. We compare the estimation of the evidence using the variational energy in Figure \ref{suff_stat} $b)$ and $d)$. We find that while our estimate using the star approximation is a much closer approximation, it does not provide a lower bound.
%

\section{Cluster variational structure learning for CTBNs}
For structure learning tasks, knowing the exact parameters of a model is in general unnecessary. For this reason we will derive an analogous but parameter-free formulation of the variational approximation for evidence and the latent state dynamics, analogous to the ones in the previous section.

\subsection{Variational structure score for CTBNs}
In the following we derive an approximate CTBN structure score, for which we need to marginalize over the parameters of the variational energy. To this end, we assume that the parameters of the CTBN are random variables distributed according to a product of local and independent \emph{Gamma distributions} $P(\mathcal{R}\mid\boldsymbol{{\alpha}},\boldsymbol{{\beta}},\mathcal{G})=\prod_{n}\prod_{x,u}\prod_{y\neq x}Gam\left[{\mathcal{R}}_{n}^u(x,y)\mid \alpha^u_n(x,y),\beta^u_n(x)\right]$ given a graph structure $\mathcal{G}$.
With our cluster approximation, the evidence is approximately given by $P(\bold{Y}\mid \mathcal{R},\mathcal{G})\approx\exp(\mathcal{{F}}_{S})$. 
By a simple analytical integration we recover an approximation to the CTBN structure score
\begin{align}
\notag&P(\mathcal{G}\mid \bold{Y},\boldsymbol{\alpha},\boldsymbol{\beta})\approx P(\mathcal{G})\int_{0}^{\infty}\mathrm{d}\mathcal{R} \ e^{\mathcal{{F}}_{S}}P(\mathcal{R}\mid\boldsymbol{\alpha},\boldsymbol{\beta},\mathcal{G})\\
&\propto e^{H}\prod_{n}\prod_{x,u}\prod_{y\neq x}{\left(\frac{\beta_n^u(x)}{(\mathbb{E}[T_n^u(x)]+\beta_n^u(x))^{M^u_n(x,y)}}\right)}^{\alpha^u_n(x,y)}\frac{\Gamma\left({\mathbb{E}[M^u_n(x,y)]+\alpha^n_n(x,y)}\right)}{\Gamma\left(\alpha^u_n(x,y)\right)},\label{score}
\end{align} 
with $\Gamma(\bullet)$ being the Gamma-function.
The approximated CTBN structure score still satisfies structural modularity, if not broken by the structure prior $P(\mathcal{G})$. However, we can not implement a \emph{k-learn} structure learning strategy as originally proposed in \cite{Nodelman2003}, as in the latent state estimation nodes become coupled and depend on each others' estimate, which in turn depend on the chosen parent set. For a detailed derivation, -- see  Appendix B.5. 
Finally we note, that in contrast to the evidence in Figure \ref{suff_stat}, we have no analytical expression for the structure score (the integral is intractable)  so that we can not compare with the exact result after integration.
%
%

\subsection{Marginal dynamics of CTBNs}
The evaluation of the approximate CTBN structure score requires the calculation of the latent state dynamics of the marginal CTBN. For this, we approximate the Gamma function in \eqref{score} via \emph{Stirling's approximation}. As Stirling's approximation becomes accurate asymptotically, we imply that sufficiently many transitions have been recorded across samples or have been introduced via a sufficiently strong prior assumption.
By extremization of the marginal variational energy, we recover a set of integro-differential equations describing the marginal self-exciting dynamics of the CTBN (see Appendix B.6). 
Surprisingly, the only difference of this parameter-free version compared to \eqref{eq:ode-back} and \eqref{eq:ode-forw} is that the conditional intensity matrix has been replaced by its posterior estimate
\begin{align}
\bar{\mathcal{R}}^{u}_n(x,y)\equiv\frac{\mathbb{E}[{M^u_n(x,y)]+\alpha^u_n(x,y)}}{\mathbb{E}[T^u_n(x)]+\beta^u_n(x)}.\label{eq:post-rm}
\end{align} 
The rate $\bar{\mathcal{R}}^{u}_n(x,y)$ is thus determined recursively by the dynamics generated by itself conditioned on the observations and prior information. We notice the similarity of our result to the one recovered in \cite{Studer2016}, where, however, the expected sufficient statistics had to be computed self-consistently during each sample path.
We employ a fixed-point iteration scheme to solve the integro-differential equation for the marginal dynamics in a manner similar to EM (for the detailed algorithm, see Appendix A.2).

\begin{table}[t!]
\caption{Experimental results with datasets generated from random CTBNs ($N=5$) with families of up to $k_{max}$ parents. To demonstrate that our score prevents over-fitting we search for families of up to $k=2$ parents. When changing one parameter the other default values are fixed to $D=10$, $b=0.6$ and $\sigma=0.2$.}
\label{tab:AU1}
\vskip 0.15in
\begin{center}
\begin{small}
\begin{sc}
\begin{tabular}{@{}lllll@{}}
\toprule
$k_{max}$ &Experiment &  Variable & AUROC & AUPR \\

\midrule
1&Number of         & $D= \: \:5$& 0.78$\pm$ 0.03&  0.64$\pm$ 0.01       \\
&Trajectories       & $D=10$& 0.87$\pm$ 0.03& 0.76$\pm$ 0.00 \\
 &                  & $D=20$&0.96$\pm$ 0.02& 0.92$\pm$ 0.00 \\
\cmidrule(l){2-5} 
&Measurement    &$\sigma=0.6$& 0.81$\pm$ 0.10&  0.71$\pm$ 0.00\\
&noise                 & $\sigma=1.0$& 0.69$\pm$ 0.07&  0.49$\pm$ 0.01\\
 \midrule

2&Number of         & $D= \: \:5$& 0.64$\pm$ 0.09&  0.50$\pm$ 0.17       \\
&Trajectories       & $D=10$&  0.68$\pm$ 0.12& 0.54$\pm$ 0.14 \\
&            & $D=20$& 0.75$\pm$ 0.11& 0.68$\pm$ 0.16 \\
\cmidrule(l){2-5} 
&Measurement    &$\sigma=0.6$& 0.71$\pm$ 0.13&  0.58$\pm$ 0.20\\
&noise                 & $\sigma=1.0$& 0.64$\pm$ 0.11&  0.53$\pm$ 0.15\\
\bottomrule
\end{tabular}
\end{sc}
\end{small}
\end{center}
\vskip -0.1in
\end{table}

\section{Results and discussion}

For the purpose of learning, we employ a \emph{greedy hill-climbing} strategy where we exhaustively score all possible families for each node with up to $k$ parents and set the highest scoring family as the current one. We do this repeatedly until our network estimate converges, which usually takes only 2 of such \emph{sweeps}. 
We can transform the scores to probabilities and generate \emph{Reciever-Operator-Characteristics} (ROCs) and \emph{Precision-Recall} (PR) curves by thresholding the averaged graphs. As a measure of performance, we calculate the averaged \emph{Area-Under-Curve} (AUC) for both. We evaluate our method using both synthetic and real-world data from molecular biology.
In order to stabilize our method in the presence of sparse data, we augment our algorithm with a prior $\boldsymbol{\alpha}=5$ and $\boldsymbol{\beta}=10$, which is uninformative of the structure, for both experiments. 
We want to stress that, while we removed the bottleneck of exponential scaling of latent state estimation of CTBNs, Bayesian structure learning via scoring still scales super-exponentially in the number of components \cite{Friedman1999}. Our method can thus not be compared to shrinkage based network inference methods such as fused graphical lasso. 

\begin{figure}[t!]
\vskip 0.2in
\begin{center}
\centerline{\includegraphics[width=0.9\columnwidth]{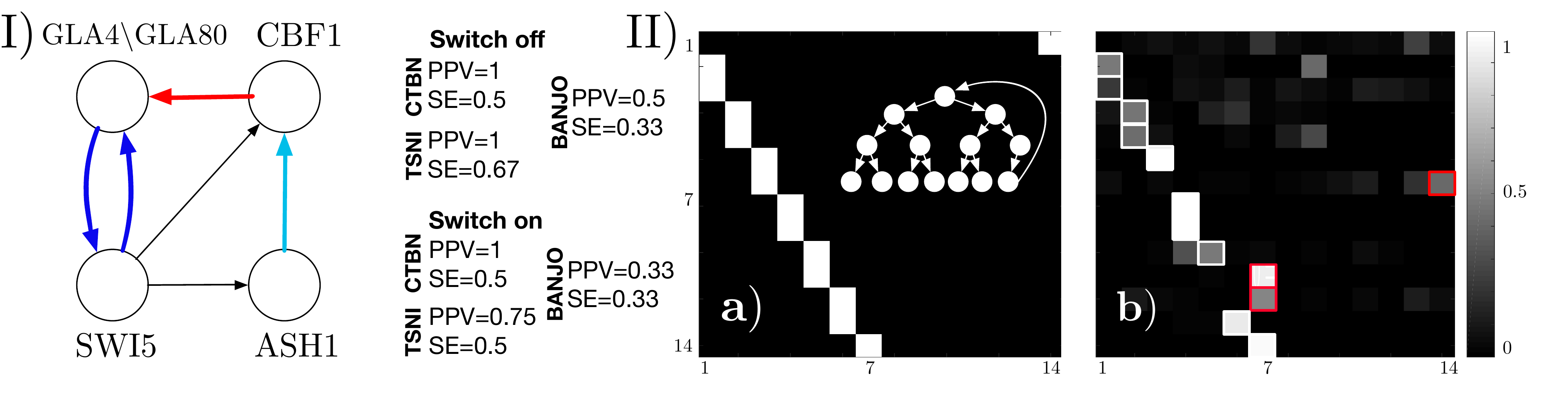}}
\caption{I) Reconstruction of a gene regulatory network (IRMA) from real-world data. To the left we show the inferred network for the "switch off" and "switch on" dataset. The ground truth network is displayed by black thin edges, the correctly inferred edges are thick (all inferred edges were correct). The the red edge was identified only  in "switch on", the teal edge only in "switch off". On the right we show a small table summarizing the reconstruction capabilities of our method, TSNI and BANJO (PPV of random guess is 0.5). II) Reconstruction of large graphs. We tested our method on a ground truth graph with 14 nodes, as displayed in $a)$ with node-relations sketched in the inlet, encoded with Glauber dynamics and searched for a maximum of $\mathrm{k}=1$ parents. 
Although we used relatively few observations that have been strongly corrupted, the averaged learned graph $b)$ is visibly close to the ground truth. We framed the prediction of the highest scoring graph, where correctly learned edges are framed white and the incorrect ones are framed red.}
\label{large_graph}
\end{center}
\vskip -0.2in
\end{figure}

The synthetic experiments are performed on CTBNs encoded with Glauber dynamics. For each of the $D$ trajectories, we recorded $10$ observations $Y^i$ at random time-points $t^i$ and corrupted them with Gaussian noise with variance $\sigma=0.6$ and zero mean.
In Table \ref{tab:AU1}, we apply our method to random graphs consisting of $N=5$ nodes, up to $k_{max}$ parents. We note that fixing  $k_{max}$ does not fix the possible degree of the node (which can go up to $N-1$). For random graphs with $k_{max}=1$ our method performs best, as expected, and we are able to reliably recover the correct graph if enough data are provided. To demonstrate that our score penalizes over-fitting we search for families of up to $k=2$ parents. 
For the more challenging scenario of $k_{max}=2$ we find a drop in performance. This can be explained by the presence of strong correlations in more connected graphs and the increased model dimension with larger $k_{max}$. In order to prove that our method outperforms existing methods in terms of scalability, we successfully learn a tree-network, with a leaf-to-root feedback, of $14$ nodes with $a=1$, $b=0.6$, see Figure \ref{large_graph} II). This is the largest inferred CTBN from incomplete data reported (in \cite{Studer2016} a CTBN of 11 nodes is learned, albeit with incomparably larger computational effort). 

Finally, we apply our algorithm to the \emph{In vivo Reverse-engineering and Modeling Assessment} (IRMA) network \cite{Cantone2009}, a synthetic gene regulatory network that has been implemented on cultures of yeast, as a benchmark for network reconstruction algorithms, see Figure \ref{large_graph} I). It is, to best of our knowledge, the only molecular biological network with a ground truth. The authors of \cite{Cantone2009} provide time course data from two perturbation experiments, referred to as "switch on" and "switch off", and attempted reconstruction using different methods. In order to compare to their results we adopt their metrics  \emph{Positive Predicted Value} (PPV) and the \emph{Sensitivity score} (SE) \cite{Bansal2007}. 
The best performing method is ODE-based (TSNI \cite{Bansal2006}) and required additional information on the perturbed genes in each experiment, which may not always be available. As can be seen in Figure \ref{large_graph} I) our method performs accurately on the "switch off" and the "switch on" data set regarding the PPV. The SE is slightly worse than for TSNI on "switch off". In both cases, we perform better than the other method based on Bayesian networks (BANJO \cite{Yu2004}). Lastly, we note that in \cite{Acerbi2014} more correct edges could be inferred using CTBNs, however with parameters tuned with respect to the ground truth to reproduce the IRMA network. For details on our processing of the IRMA data, see Appendix C.

\section{Conclusion}
We develop a novel method for learning directed graphs from incomplete and noisy data based on a continuous-time Bayesian network. To this end, we approximate the exact but intractable latent process by a simpler one using cluster variational methods. We recover a closed set of ordinary differential equations that are simple to implement using standard solvers and retain a consistent and accurate approximation of the original process. Additionally, we provide a close approximation to the evidence in the form of a variational energy that can be used for learning tasks. Lastly, we demonstrate how marginal dynamics of continuous-time Bayesian networks, which only depend on data, prior assumptions, and the underlying graph structure, can be derived by the marginalization of the variational free energy. 
Marginalization of the variational energy provides an approximate structure score. We use this to detect the best scoring graph using a greedy hill-climbing procedure. It would be beneficial to identify higher-order approximations of the variational energy in the future.  We test our method on synthetic as well as real data and show that our method produces meaningful results while outperforming existing methods in terms of scalability.


\bibliography{cv_learning_ctbn}
\bibliographystyle{plain}

\clearpage
\section*{Supplementary Material}
\appendix

 \section{Algorithms}
 In this section we give the detailed algorithms described in the main text. All equation references point to the main text. 
 \subsection{Stationary points of Euler--Lagrange equation}
 \begin{algorithm}[h]
   \caption{Stationary points of Euler--Lagrange equation}
   \label{alg:euler-lagrange}
\begin{algorithmic}[1]
   \STATE {\bfseries Input:} Legal set of initial trajectories ${m}_n(x)$, boundary conditions $m(x,0)$ and $\rho(x,T)$, \\Observations $\mathbf{Y}$.
   \REPEAT
   \FORALL {$n\in \{1,\dots,N\}$ }
   \FORALL {$Y^i\in\mathbf{Y}$} 
   \STATE Update $\rho_n(x)$ by backward propagation from $t_i$ to $t_{i-1}$ using (5) fulfilling reset conditions (6).
   \ENDFOR
    \STATE  Update $m_n(x)$ by forward propagation using (4) given $\rho_n(x)$.
    \ENDFOR
   \UNTIL{Convergence}
   \STATE {\bfseries Output:} Set of $m_n(x)$ and $\rho_n(x)$.
\end{algorithmic}
\end{algorithm}
 \subsection{Marginal CTBN dynamics}
 \begin{algorithm}[h]
   \caption{Marginal CTBN dynamics}
   \label{alg:marg-dynamics}
\begin{algorithmic}[1]
   \STATE {\bfseries Input:} 
   Propose set of initial trajectories ${m}_n(x)$, \\observations $\bold{Y}$, prior assumption on sufficient statistics $\boldsymbol{\alpha}$ and $\boldsymbol{\beta}$,
   initial guess for $\bar{\mathcal{R}}^{u}_n(x,y)$.
   \REPEAT 
   \STATE Set current  $\bar{\mathcal{R}}^{u}_n(x,y)$ as current CIM.   
   \STATE {Solve marginal dynamic equation with $\bar{\mathcal{R}}^{u}_n(x,y)$ using Algorithm\ref{alg:euler-lagrange}.}
   \STATE {Use expected sufficient statistics to update $\bar{\mathcal{R}}^{u}_n(x,y)$ via (8). }
   \UNTIL{Convergence}
   \STATE {\bfseries Output:} Set of $m_n(x)$ and $\rho_n(x)$.
\end{algorithmic}
\end{algorithm}

\section{Derivations}

 \subsection{Variational energy in star approximation}\label{app:A}
 
 In the following we derive the star approximation of a factorized stochastic process. In order to lighten the notation we omit the corresponding process to each variable from now on, with $X(t+\delta t)=y$, $X(t)=x$. The exact expression of the variational energy $\mathcal{F}[Q]$ for a continuous-time Markov process decomposes into time-wise energies $\mathcal{F}[Q]=\sum_t f(t)$ with
 \begin{align*}
f(t)&\equiv\underset{\equiv{\langle H(t) \rangle}_Q}{\underbrace{\int_{\mathcal{X}}\mathrm{d}x\,\int_{\mathcal{X}}\mathrm{d}y\,Q(y\mid x)Q(x)\ln P(y\mid x)}}-\underset{\equiv S[Q](t)}{\underbrace{\int_{\mathcal{X}}\mathrm{d}x\,\int_{\mathcal{X}}\mathrm{d}y\,Q(y\mid x)Q(x)\ln Q(y\mid x)}},
\end{align*}
where we identified the time-dependent energy function $H(x,y,t) \equiv \ln P(y\mid x)$ and the entropy $S[Q](t)$.
In the following, we explicitly use $x=(x_1,\dots,x_N)$ and $y=(y_1,\dots,y_N)$ for $x,y\in\mathcal{X}$ with $x_n\in\mathcal{X}_n$ and  $y_n\in\mathcal{X}_n$.
We assume $Q(\bold{X})$ to describe a factorized stochastic process, i.e. $Q(y,x)=\prod_n Q(y_n\mid x_n,u_n)Q( x)$, where we introduced the process of neighbours $U_n(t)=u_n$.
Consider now 
\begin{align*}
{\langle H(t) \rangle}_Q&= \int\int_{\mathcal{X}}\mathrm{d}x\,\mathrm{d}y\,Q(x)\prod_nQ(y_n\mid x_n, u_n)\ln\prod_kP(y_k\mid x_k, u_k)\\
\end{align*}
Assuming  temporal correlations with neighboring nodes scale with $\varepsilon$
\begin{align}
P(y_n\mid x_n, u_n)\equiv P(y_n\mid x_n)+\mathcal{O}(\varepsilon),\label{eq:expansion-explicit}
\end{align}
where the remainder $\mathcal{O}(\varepsilon)$ contains dependency on the parents. Naturally, we assume that the approximating distribution $Q$ has a similar expansion in the same parameter $\varepsilon$. We get the approximate entropy using Appendix  \ref{Expansion formula I} in first order of $\varepsilon$
\begin{align*}
{\langle H(t) \rangle}_Q=&\int\int_{\mathcal{X}}\mathrm{d}x\,\mathrm{d}y\,Q(x)\left[\sum_nQ(y_n\mid x_n, u_n)\prod_{m\neq n}Q(y_m\mid x_m)-(N-1)\prod_{m}Q(y_m\mid x_m)\right]\sum_k\ln P(y_k\mid x_k, u_k).
\end{align*}
For $n\neq k$ we can sum over $y_n$.
This leaves us with
\begin{align*}
{\langle H(t) \rangle}_Q=&\int\int_{\mathcal{X}}\mathrm{d}x\,\mathrm{d}y\,Q(x)\sum_nQ(y_n\mid x_n,u_n)\ln P(y_n\mid x_n, u_n)+\mathcal{O}(\varepsilon^2).
\end{align*}
The exact same treatment can be done for the entropy term and we arrive at variational free energy in \emph{star-approximation}
\begin{align*}
\mathcal{F}[Q]=&\sum_{t,n} \int\int_{\mathcal{X}_n}\mathrm{d}x_n\,\mathrm{d}y_n\,Q(y_n,x_n,u_n)
\left[\ln P(y_n\mid x_n, u_n)-\ln Q(y_n \mid  x_n,u_n)\right]+\mathcal{O}(\varepsilon^2)
\end{align*}

\subsubsection{Expansion formula I} \label{Expansion formula I}
 Note that for $\prod_{n=1}^N Q_n$ with $Q_n=a_{n}+\varepsilon b_{n}$ for any $a_n,b_n\in\mathbb{R}$  holds
 \begin{align*}
 \prod_{n=1}^N  Q_n=\sum_{m=1}^{N}Q_{m}\prod_{{n\neq m},n=1}^N a_{n}-(N-1) \prod_{n=1}^Na_{n}+\mathcal{O}(\varepsilon^2)
 \end{align*}
 Proof:
 \begin{align*}
\prod_{n=1}^N Q_{n}&=\prod_{n=1}^N a_{n}+\epsilon\sum_{m=1}^{N}b_{m}\prod_{{n\neq m},n=1}^Na_{n}
+\sum_{m=1}^{N}a_{m}\prod_{{n\neq m},n=1}^Na_{n}-\sum_{m=1}^{N}a_{m}\prod_{{n\neq m},n=1}^Na_{n}+\mathcal{O}(\varepsilon^2)\\
=&\sum_{m=1}^{N}\left[a_{m}+\epsilon b_{m}\right]\prod_{{n\neq m},n=1}^N a_{n}+\prod_{n=1}^N a_{n}-\sum_{m=1}^{N}\prod_{n=1}^Na_{n}+\mathcal{O}(\varepsilon^2)\\
=&\sum_{m=1}^{N}Q_{m}\prod_{{n\neq m},n=1}^N a_{n}-(N-1)\prod_{n=1}^Na_{n}+\mathcal{O}(\varepsilon^2).
 \end{align*}

 
 \subsection{Continuous-time variational energy in star approximation}\label{app:B}
\label{app:Cluster factorization}
In order to perform the continuous-time limit, we represent $Q$ by an expansion in $\delta t$ in set of marginals
 \begin{align*}
&Q(y_n,x_n,u_n)=\delta(x,y){m}_n(x)m_n^u+\tau^u_n(x,y)\delta t+{o}(\delta t),
\end{align*}
with $\tau^u_n(x,x)=-\sum_{y\neq x} \tau^u_n(x,y)$. 
By inserting $Q's$ representation  into $\mathcal{F}$ we get
\begin{align*}
\mathcal{{F}}=-&\sum_{n}\sum_{t}\sum_{x,y\neq x,u}\delta t\tau_{n}^{u}(x,y)\left[\ln\delta t\frac{\tau_{n}^{u}(x,y)}{m_n(x)m_{n}^{u}}-\ln\delta t\mathcal{R}_{n}^{u}(x,y)\right]\\
-&\sum_{n}\sum_{t}\sum_{x,u}\left[m_{n}^{u}m_{n}(x)-\delta t\sum_{y\neq x}\tau_{n}^{u}(x,y)\right]\times\left[\ln[1-\frac{\delta t\sum_{y\neq x}\tau_{n}^{u}(x,y)}{m_n(x)m_{n}^{u}}]-\ln[1+\mathcal{R}_{n}^{u}(x,x)\delta t]\right]
\end{align*}
where we also inserted  $P( X_n(t)=y_n\mid X_n(t)=x_n,U_n(t)=u)=\delta_{x,y}+\mathcal{R}_n^u(x,y)\delta t$. With the asymptotic identity $\ln (1+\delta tx)= \delta tx$ we can simplify
\begin{align*}
\mathcal{{F}}=-&\sum_{n}\sum_{t}\sum_{x,y\neq x,u}\delta t\tau_{n}^{u}(x,y)\left[\ln \frac{\tau_{n}^{u}(x,y)}{m_n(x)m_{n}^{u}}-\ln \mathcal{R}_{n}^{u}(x,y)\right]\\
+&\sum_{n}\sum_{t}\sum_{x,u}\left[m_{n}^{u}m_{n}(x)-\delta t\sum_{y\neq x}\tau_{n}^{u}(x,y)\right]\times\left[\frac{\delta t\sum_{y\neq x}\tau_{n}^{u}(x,y)}{m_n(x)m_{n}^{u}}+\mathcal{R}_{n}^{u}(x,x)\delta t]\right]
\end{align*}
which becomes in the continuous-time limit $\delta t \rightarrow 0$ 
\begin{align*}
\mathcal{{F}}=&\sum_{n}\int\mathrm{{d}}t\sum_{x,y\neq x,u}\tau_{n}^{u}(x,y)[1-\ln\tau_{n}^{u}(x,y)+\ln(m_{n}^{u}m_{n}(x))]\\
+&\sum_{n}\int\mathrm{{d}}t\left[\sum_{x,u}m_{n}(x)m_{n}^{u}\mathcal{R}_{n}^{u}(x,x)+\sum_{x,y\neq x,u}\tau_{n}^{u}(x,y)\ln\mathcal{R}_{n}^{u}(x,y)\right].
\end{align*}
The contribution of the likelihood term can be derived to be
\begin{align*}
\mathcal{F}_0&=\sum_t \delta t \sum_{i}\mathbb{E}_{N}[\ln P(Y^i\mid x)]\frac{\delta(t,t^i)}{\delta t}\underrel{\delta t\to 0}{=}  \int_0^T\mathrm{d}t \sum_{i}\mathbb{E}_{N}[\ln P(Y^i\mid x)]\delta(t-t^i),\\
\mathbb{E}_{N}[f(x)]&=\sum_{x\in\mathcal{X}}f(x)\prod_{k=1}^N m_k(x_k), \quad x_k\in\mathcal{X}_k.
\end{align*}
 \subsection{Naive mean-field approximation}\label{app:C}
We recover the variational energy in naive mean-field approximation by only consider the zeroth order expansion in the correlations $\varepsilon$, meaning
\begin{align*}
Q(y_n\mid x_n,u_n)=Q(y_n\mid x_n)
\end{align*}
Then for the entropy $S(t)$ from Appendix \ref{app:A} holds
\begin{align*}
S(t)&\equiv \sum_{y,x}Q(y,x)\ln Q(y,x)=  \sum_{y,x}\prod_n Q(y_n,x_n)\ln\left[\prod_m Q(y_m,x_m)\right]\\
\end{align*}
Thus for the variational energy we arrive at the naive mean-field approximation
\begin{align*}
\mathcal{{F}}=&\sum_{n}\sum_t \sum_{y_n,x_n}Q(y_n,x_n)\sum_{u}\prod_{k\in\mathrm{pa}(n)}Q(u_k)\ln P(y_n\mid x_n,u_n)-\sum_{n}\sum_t \sum_{y_n,x_n}Q(y_n,x_n)\ln Q(y_n,x_n)\\
&+\sum_{n}\sum_t\sum_{x_n}Q(x_n)\ln Q(x_n)
\end{align*}
Finally considering the marginals of the transitions
\begin{align*}
Q(y_n,x_n)=\delta_{x,y}m_n(x)+\tau_n(x,y)+o(\delta t),
\end{align*}
we recover the result from (Cohn, El-Hay, Friedmann and Kupfermann 2010) using an identical derivation as given in Appendix \ref{app:B}.

\subsection{CTBN dynamics in star approximation}\label{app:D}
\label{app:Euler-Lagrange}
We are now going to derive the dynamics of CTBNs in star approximation, defined by fulfilling the Euler--Lagrange equations 
\begin{align*}
\partial_{x}\mathcal{{L}}[t,x,\dot{{x}}]-\partial_{t}[\partial_{\dot{{x}}}\mathcal{{L}}[t,x,\dot{{x}}]]=0.
\end{align*}
First lets consider the derivative with respect to $m_n(x)$:
\begin{align*}
\partial_{m_n(x)}H_n=\sum_{u}\sum_{y\neq x}\frac{\tau_n^u(x,y)}{m_n(x)},\quad
\partial_{m_n(x)}E_j=\mathbb{E}_{n}[\mathcal{R}_n^u(x,x)],
\end{align*}
Further if node $n$ has a child $j$
\begin{align*}
\partial_{m_n(x)}H_j=\sum_{x,u\mid X_n(t)=x_n=x}\sum_{y\neq x}\frac{\tau^u_j(x,y)}{m_n(x)},\quad
\partial_{m_n(x)}E_j=\sum_{x}m_j(x)\mathbb{E}_n[\mathcal{R}_n^u(x,x)\mid X_n(t)=x].
\end{align*}
With respect to the derivative $\dot{m}_n(x)$ we get
\begin{align*}
\partial_{\dot{m}_n(x)}\mathcal{{L}}=-\lambda_n(x).
\end{align*}
We derive with respect to the transitions 
\begin{align*}
\partial_{\tau_n^u(x,y)}H_n=\ln [m_n(x)m_n^u]-\ln \tau_n^u(x,y),\quad
\partial_{\tau_n^u(x,y)}E_n=\ln  \mathcal{R}_n^u(x,y).
\end{align*}
thus
\begin{align*}
\partial_{\tau_n^u(x,y)}\mathcal{L}=\ln [m_n(x)m_n^u]-\ln \tau_n^u(x,y)+\ln \mathcal{R}_n^u(x,y)-\lambda_n(x)+\lambda_n(y).
\end{align*}
The derivative with respect to the Lagrange-multipliers yields:
\begin{align*}
\partial_{\lambda_n(x)}\mathcal{L}=-\left\{\dot{m}_n(x)-\left[\sum_{y\neq x,u}\tau_n^u(y,x)-\tau_n^u(x,y)\right]\right\}
\end{align*}
And lastly assuming a factorized noise model $P(Y^i|X(t)=x)=\prod_n P_n(Y^i|X_n(t)=x_n)$ we have for the derivative of $\mathcal{F}_0$
\begin{align*}
\partial_{m_n(x)}\mathcal{F}_0=\sum_i \ln P_n(Y^i|x)\delta(t-t^i)
\end{align*}

These can then be combined as the following Euler-Lagrange equations:
\begin{align*}
&(\mathrm{I})\quad0=\sum_{u}\sum_{y\neq x}\frac{\tau_n^u(x,y)}{m_n(x)}+\mathbb{E}_{n}[\mathcal{R}_n^u(x,y)]+\dot{\lambda}_n(x)+\sum_i \ln P_n(Y^i|x)\delta(t-t^i)\\
&\quad\quad\quad+\sum_{j\in \mathrm{child}(n)}\sum_{x,u\mid X_n(t)=x}\sum_{y\neq x}\frac{\tau_n^u(x,y)}{m_n(x)}+\sum_{x}m_j(x)\mathbb{E}_{j}[r^u_j(x,x)\mid X_n(t)=x]\\
&(\mathrm{II})\quad0=\ln [m_n(x)m_n^u]-\ln \tau_n^u(x,y)+\ln \mathcal{R}_n^u(x,y)-\lambda_n(x)+\lambda_n(y)\\
&(\mathrm{III})\quad\dot{m}_n(x)=\sum_{y\neq x,u}\tau_n^u(y,x)-\tau_n^u(x,y).
\end{align*}
Exponentiating $(\mathrm{II})$ gives
\begin{align*}
(\mathrm{II}^*)\quad\tau_n^u(x,y)=m_n(x)m_n^u \mathcal{R}_n^u(x,y)\rho_n(y)/\rho_n(x),
\end{align*}
where $\rho_n(x)\equiv\exp(\lambda_n(x))$. Assuming that $\mathcal{R}$ is irreducible, $\rho_n(x)$ and $m_n(x)$ are non-zero in $(0,T)$
 and we can thus eliminate $\tau_n^u(x,y)$ in $(\mathrm{I})$ and $(\mathrm{II})$.
 Thus
 \begin{align*}
&(\mathrm{I}^*)\quad\dot{\rho}_n(x)=\sum_{y\neq x}\mathbb{E}_{n}[\mathcal{R}_n^u(x,y)]\rho_n(y)+\left\{\mathbb{E}_{n}[\mathcal{R}_n^u(x,x)]+\psi_n\right\}\rho_n(x)\\
&(\mathrm{III}^*)\quad\dot{m}_n(x)=\sum_{y\neq x}\left\{m_n(y)\mathbb{E}_{n}[\mathcal{R}_n^u(y,x)]\rho_n(x)/\rho_n(y)-m_n(x)\mathbb{E}_{n}[\mathcal{R}_n^u(x,y)]\rho_n(y)/\rho_n(x)\right\},
 \end{align*}
 where we used $\dot{\lambda}_n(x)=\frac{1}{\rho_n(x)}\dot{\rho}_n(x)$. We further summarized
 \begin{align*}
 \psi_n=&\sum_{j\in\mathrm{child}(n)}\sum_{x}m_j(x)\left\{ \sum_{y\neq x}\frac{\rho_j(y)}{\rho_j(x)}\mathbb{E}_{n}[r^u_j(x,y)\mid X_n(t)=x]+\mathbb{E}_{n}[r^u_j(x,x)\mid X_n(t)=x]\right\}\\
 &+\sum_i \ln P_n(Y^i|x)\delta(t-t^i).
 \end{align*}
 The driving term $\ln P_n(Y^i|x)\delta(t-t^i)$ then conditions the dynamics on the observations by
 $\lim_{t\rightarrow t^{i-}}\rho_n(x)=\lim_{t\rightarrow t^{i+}} P_n(Y^i|x)\rho_n(x)$. 
 \subsection{Variational marginal score}\label{app:E}
 Using $\mathcal{R}_n^u(x,x)=-\sum_{y\neq x}\mathcal{R}_n^u(x,y)$ we can write
 \begin{align*}
 E_{n}=&\int dt\sum_{x,u}\sum_{y\neq x} \left[\tau_n^u(x,y)\ln \mathcal{R}_n^u(x,y)-m_n(x)m_n^u \mathcal{R}_n^u(x,y)\right].
 \end{align*}
 For the approximated evidence
 \begin{align*}
 P(\bold{Y}\mid\mathcal{{R}})\approx\prod_{n}\exp\left[E_{n}+H_{n}\right].
 \end{align*}
 we get 
  \begin{align*}
 P(\bold{Y}\mid\mathcal{{R}})\approx e^{H}\prod_{n}\prod_{x,u}\prod_{y\neq x}\mathcal{R}_n^u(x,y)^{\mathbb{E}[M_n^u(x,y)]} e^{-\mathbb{E}[T_n^u(x)] \mathcal{R}_n^u(x,y)}.
 \end{align*}
 Assuming an independent \emph{Gamma} prior
  \begin{align*}
P(\mathcal{R}\mid\boldsymbol{{\alpha}},\boldsymbol{{\beta}},\mathcal{G})=&\prod_{n}\prod_{x,u}\prod_{y\neq x}\gamma\left[{r}^u_n(x,y)\mid \alpha^u_n(x,y),\beta^u_n(x)\right]\\
=&\prod_{n}\prod_{x,u}\prod_{y\neq x}\frac{{\beta_n^u(x)}^{\alpha_n^u(x,y)}}{\Gamma(\alpha_n^u(x,y))}{\mathcal{R}_n^u(x,y)}^{\alpha_n^u(x,y)-1}e^{-\beta_n^u(x) \mathcal{R}_n^u(x,y)}.
 \end{align*}
 Thus we can express the graph posterior
  \begin{align*}
 P(\mathcal{{G}}|\bold{Y},\boldsymbol{{\alpha}},\boldsymbol{{\beta}})\propto& P(\mathcal{G})\int_{0}^{\infty}P(\bold{Y}|\mathcal{{R}})P(\mathcal{{R}}|\boldsymbol{{\alpha}},\boldsymbol{{\beta}},\mathcal{G})\ \mathrm{d}{\mathcal{{R}}}\\
 =&e^{H}\prod_{n}\prod_{x,u}\prod_{y\neq x}\frac{{\beta_n^u(x)}^{\alpha_n^u(x,y)}}{\Gamma(\alpha_n^u(x,y))}\\
 &\times \int_{0}^{\infty} {\mathcal{R}_n^u(x,y)}^{\mathbb{E}[M_n^u(x,y)]+\alpha_n^u(x,y)-1} e^{-[\mathbb{E}[T_n^u(x)]+\beta_n^u(x)] \mathcal{R}_n^u(x,y)}  \mathrm{d}\mathcal{R}_n^u(x,y),
  \end{align*}
 which has an analytic solution
\begin{align*}
\int_{0}^{\infty}x^{a}e^{-bx}\mathrm{{d}}x=b^{-a}\Gamma(a).
\end{align*}
Thus we get
\begin{align*}
P(\mathcal{{G}}|\bold{Y},\boldsymbol{{\alpha}},\boldsymbol{{\beta}})\propto e^{H}\prod_{n}\prod_{x,u}\prod_{y\neq x}\left(\frac{{{\beta_n^u(x)}}}{(\mathbb{E}[T_n^u(x)]+\beta_n^u(x))^{\mathbb{E}[M_n^u(x,y)]}}\right)^{\alpha_n^u(x,y)}\frac{\Gamma(\mathbb{E}[M_n^u(x,y)]+\alpha_n^u(x,y))}{\Gamma(\alpha_n^u(x,y))}.
\end{align*}
 
 \subsection{Marginal dynamics for CTBNs}\label{app:F}
In the following we are going to derive the dynamic equations of the marginal process for which we have to expand the Gamma-function.
Assuming the sum of recorded transitions and prior transition number to be sufficiently large we can approximate the Gamma-function using Stirling's approximation $\Gamma(z)\approx\sqrt{{\frac{{2\pi}}{z}}}\left(\frac{{z}}{e}\right)^{z}+\mathcal{{O}}\left(\frac{{1}}{z}\right)$ we get the approximate marginal score function
\begin{align*}
\ln P(\mathcal{{G}}|\bold{Y},\boldsymbol{{\alpha}},\boldsymbol{{\beta}})\propto\sum_{n}H_{n}+\mathcal{{E}}_{n},
\end{align*}
with 
\begin{align*}
\mathcal{{E}}_{n}=&\sum_{x,u}\sum_{y\neq x}\left(\mathbb{E}[M_n^u(x,y)]+\alpha_n^u(x,y)-\frac{{1}}{2}\right)\ln\left(\mathbb{E}[M_n^u(x,y)]+\alpha_n^u(x,y)\right)\\
&-\left(\alpha_n^u(x,y)-\frac{{1}}{2}\right)\ln\left(\alpha_n^u(x,y)\right)+\alpha_n^u(x,y)\ln\left(\beta_n^u(x)\right)\\
&-\left(\mathbb{E}[M_n^u(x,y)]+\alpha_n^u(x,y)\right)\ln\left(\mathbb{E}[T_n^u(x)]+\beta_n^u(x)\right)-\mathbb{E}[M_n^u(x,y)],
\end{align*}
 \begin{align*}
 &\partial_{m_n(x)}{\mathcal{{E}}_{n}}=-\sum_{y\neq x}\sum_{u}\mathbb{E}_n\left[\frac{\mathbb{E}[M_n^u(x,y)]+\alpha_n^u(x,y)}{\mathbb{E}[T_n^u(x)]+\beta_n^u(x)}\right],\\
 &\partial_{m_n(x)}{\mathcal{{E}}_{j}}=-\sum_{x}\sum_{y\neq x} m_j(x)\mathbb{E}_n\left[\frac{\mathbb{E}[M_j^u(x,y)]+\alpha_j^u(x,y)}{\mathbb{E}[T_j^u(x)]+\beta_j^u(x)}\mid X_n(t)=x\right]\mathbbm{1}[j\in\mathrm{child}(n)],\\
 &\partial_{\tau_n^u(x,y)}\mathcal{{E}}_{n}\approx\ln\left(\frac{\mathbb{E}[M_n^u(x,y)]+\alpha_n^u(x,y)}{\mathbb{E}[T_n^u(x)]+\beta_n^u(x)}\right),
 \end{align*}
 where we approximated $\frac{\mathbb{E}[M_n^u(x,y)]+\alpha_n^u(x,y)-\frac{{1}}{2}}{\mathbb{E}[M_n^u(x,y)]+\alpha_n^u(x,y)}-1\approx 0$. The derivatives with respect to $H_n$ and the constraint remain unchanged, see Appendix \ref{app:D}.
Finally defining the posterior-rate
\begin{align*}
\bar{\mathcal{R}}^u_n(x,y)\equiv \frac{\mathbb{E}[M_n^u(x,y)]+\alpha_n^u(x,y)}{\mathbb{E}[T_n^u(x)]+\beta_n^u(x)},
\end{align*}
 we arrive at the same set of equations as in Appendix \ref{app:D}.

\section{Processing IRMA data}
In this section we present our approach of processing IRMA data. The IRMA dataset consists of expression data of genes, measured in concentrations, which are continuous. We can not capture continuous data using CTBNs, but need to map this data to a set of latent states. We identify two states \emph{over-expressed} ($X=1$) and  \emph{under-expressed} ($X=0$) with respect to the \emph{basal} (equilibrium) concentration $c_B$. This motivates the following observation model given the basal concentration
\begin{align}
&P(Y\mid X=1,c_B)=
\begin{cases}
1/|Y_0|&, Y \geq c_B \text{ and } Y \leq Y_0\\
0&, else
\end{cases},\\
&P(Y\mid X=0,c_B)=\begin{cases}
1/|Y_0|&, Y < c_B \text{ and }  Y \geq -Y_0\\
0&, else
\end{cases},
\end{align}
where we have to choose some $Y_{0}$, so that the likelihood is normalized. We set $Y_0$ to some large value $Y_0\geq \mathrm{argmax}_{|Y|\in \mathrm{DATA}}$ as our method remains invariant under each choice.

We model the basal concentration itself is a random variable, which we assume is gaussian distributed. We can estimate the parameters of the gaussian distribution $\mu_B$ and $\sigma_B$ from the data. The marginal observation model is then acquired by integration
\begin{align}
P(Y\mid X)=\begin{cases}
1-\mathrm{erf}((Y-\mu_B)/\sigma_B)&, X=1\\
\mathrm{erf}((Y-\mu_B)/\sigma_B)&, X=0
\end{cases}.
\end{align}
Given this observation model we can assign each measurement a likelihood and can process the data using our method. We note that other models for IRMA data can be thought of that may return better (or worse) results using our method.

\end{document}